\documentclass[runningheads]{llncs}
\usepackage[T1]{fontenc}
\usepackage{graphicx}
\usepackage{amsmath}

\begin{document}
\title{The Effects of Population Size on the Performance of BEAGLE GPU-Based Genetic Programming Runs}
\titlerunning{The Effects of Population Size on GPU-based BEAGLE GP Runs}
%
\author{Nathan Haut\inst{1}
\and
Ilya Basin\inst{2}
\and
Ruchika Gupta\inst{1}
\and
Marzieh Kianinejad\inst{1}
\and
Zachary Perrico\inst{1}
\and
Elijah Smith\inst{1}
\and
Wolfgang Banzhaf\inst{1}
}
\authorrunning{N. Haut et al.}
%
\institute{Michigan State University, East Lansing MI 48824, USA, 
\email{hautnath@msu.edu}\\
\and
Noblis, Reston VA , USA,
\email{Ilya.Basin@noblis.org}}
\maketitle              
\begin{abstract}
The Beagle framework, through GPU-based Genetic Programming, enables population dynamics previously unattainable (within practical time frames) by CPU-constrained Genetic Programming systems. This work explores how GPU-enabled population sizes  impact the success of training for symbolic regression problems. Specifically, when using constant population sizes, we see benefits of using very narrow and deep searches (as narrow as 1000 individuals) for some problems, while other problems benefit from very broad and shallow searches (as broad as 10 million individuals). We also explore stepped population sizes that start with large populations and drop to small populations to balance the breadth and depth of search. 


\end{abstract}
%


%

\section{Introduction}

In recent years, the widespread adoption of Graphics Processing Units (GPUs) \cite{oh2004gpu,steinkrau2005using,pandey2022transformational} has revolutionized computationally intensive fields by providing massive parallel processing capabilities. However, the impact of GPU computing on Evolutionary Computation has historically lagged behind its transformative effect on Artificial Neural Networks (ANNs). While the deep learning boom was fueled by software frameworks that made matrix calculations easy for GPUs, evolutionary algorithms lacked this underlying infrastructure, creating massive engineering hurdles for researchers trying to use GPUs for parallel population processing. \cite{lange2023neuroevobench}.

Genetic Programming (GP) \cite{koza1992genetic}, a subset of Evolutionary Algorithms (EA), is uniquely positioned to benefit from this hardware. As a heuristic search technique, GP mimics the principles of natural selection to search for global optima by iteratively optimizing a population of computer programs, relying on a fitness function to evaluate and guide each program across every data point in a training set. This evaluation phase creates a massive computational bottleneck that consumes the vast majority of execution time.

Although early efforts to accelerate Genetic Programming (GP) via GPUs successfully reduced evaluation times \cite{harding2007fast,ferigo2020gpu}, their ultimate performance remained constrained by their reliance on traditional Tree-Based GP representations \cite{robilliard2009high,zhang2022speeding}. Since GPUs are optimized for linear contiguous memory operations, the inherent challenge of mapping irregular dynamic tree structures onto a rigid Single Instruction, Multiple Data (SIMD) architecture has forced researchers to rely on a variety of complex workarounds in recent literature \cite{sathia2021accelerating,zhang2022speeding}. While the first end-to-end Genetic Algorithms were hosted on GPUs as early as 2005 \cite{wong2005parallel,fok2007evolutionary}, these pioneering systems were forced to omit complex operations like crossover due to their expensive overhead of structural recombination. 

To bridge this gap, modern solutions typically rely on serializing trees into prefix arrays or Reverse Polish Notation (RPN) \cite{fan2026asynchronous,robilliard2009high,sathia2021accelerating}. Other approaches attempted to use high-level wrappers like TensorGP\cite{baeta2021tensorgp} and KarooGP\cite{staats2017tensorflow}, two common GPU-enabled GP frameworks that support symbolic regression. However, they suffer from inflexible dataset constraints or slow graph execution models \cite{zhang2022speeding}. To satisfy the tensor requirements of modern AI frameworks like PyTorch, frameworks such as EvoGP \cite{wang2025evogp} pad smaller trees with empty "NaN" values up to a maximum length. While this enables batch processing, it inherently causes severe memory bloat and computational waste, as the GPU spends valuable cycles processing empty padding.

Beyond the structural challenges of trees, the limited impact of GPUs on GP is rooted in an 'occupancy' problem. GPUs are designed to run extremely fast by having thousands of threads execute the exact same instruction at the exact same time \cite{langdon2011graphics}. However, when threads encounter conditional branches (like IF/ELSE statements), warp divergence occurs. To counter this, researchers like Arora et al. \cite{arora2010parallelization} advocated for 'algebraic masking'—replacing logical IF branches with complex mathematical expressions that evaluate all execution paths simultaneously. While this 'Arora trick' maintains high occupancy by preventing thread stalls, it does so by forcing the GPU to perform redundant calculations for every branch. Vimarsh et. al \cite{sathia2021accelerating} counteracted this by forcing every thread-block to evaluate the exact same program. However, their approach relies on Arrays of Structures (AoS) with nested pointers, which results in memory transfer bottlenecks delaying the training time.

Consequently, scaling GP on GPUs has historically resulted in
a compromise of hardware idleness of divergence, the execution bloat of structural node-parsing, and/or memory transfer bottlenecks. Yet, as noted by Lange et al. \cite{lange2023neuroevobench}, the need to evaluate GP against modern neural architectures makes this the critical time to reassess evolutionary optimization at massive scales—provided that these deep-rooted hardware incompatibilities can finally be overcome.

To answer this challenge, we introduce Beagle—an open-source symbolic regression framework developed by Ilya Basin at Noblis, Inc. that resolves the historical bottlenecks of warp divergence and Garbage Collection (GC). This allows us to unlock and subsequently explore  population sizes in the millions of individuals. 

The key contributions in this work are: 

\begin{itemize}\item \textbf{Quantifying Scaling Limits}: We evaluate the impact of massive population breadths at different generational depths achievable through Beagle’s GPU-accelerated architecture.

\item \textbf{Optimization Trade-offs}: We characterize the relationship between population breadth and generational depth across symbolic regression tasks of varying complexity, given that tradeoff between the two. 

\item \textbf{Dynamic Population Management}: We investigate starting with greater population breath and reducing it after a certain number of generations to achieve a greater generational depth in order to benefit from both. This approach enables amplifying selection pressure "mid-flight" and increasing  the speed of progress through generations.\end{itemize}

\section{Related Work}

Selecting the right population size has been essential to successful results in genetic programming since the method's inception. Dynamically adjusting population size across generations then, either to search for an efficient size or to encourage specific behavior, is a natural extension of this understanding.

With the goal of reducing computational effort, Tomassini et al~\cite{tomassini2004new} explored a "plague" technique, wherein individuals were removed when the algorithm was progressing smoothly and added when the algorithm was stagnating. Kouchakpour et al~\cite{kouchakpour2009dynamic} improved this technique with a "pivot function" that assessed the population's fitness improvement at every generation. If the output of this function exceeded the user-defined "pivot point," then the population size was reduced; population size was increased if the opposite occurred. Nine years later, Tao et al~\cite{tao2013new} further expanded the technique with an exponential pivot function (EXP) that produced more efficient solutions than previous pivot methods.

Genetic Programming's similarity to Genetic Algorithms (GAs) allows some dynamic size strategies to be cross-compatible. Hu and Banzhaf~\cite{hu2009role} observed several schemes that could cross over from GAs to GP, and identified the rate of genetic insertions as a new indicator for population size adjustment. Hu et al~\cite{hu2010variable} examined the effects of dynamic size strategies in GP when used in tandem with parallel processing on a GPU and found that these combined strategies could accelerate evolution. Since Beagle also uses GPU parallelization, the results of this work are particularly relevant. Finally, a recent survey by Farrenati and Vanneschi~\cite{farinati2024survey} documented the efficiency improvements of dynamic size strategies over static size strategies in numerous bio-inspired algorithms. These included Evolutionary Algorithms such as GP, Particle Swarm Optimization, Ant Colony Optimization, and Artificial Bee Colony algorithms.

The Beagle framework described in this work utilizes GPUs (using specifically NVIDIA's CUDA software library), as has been previously mentioned; therefore, we also examine previous applications of GPUs in genetic programming. In one of their first applications to a GP problem, Harding and Banzhaf~\cite{harding2007fast} showed that GPUs could perform several hundred times faster than CPUs when evaluating a Cartesian-GP individual if the number of fitness cases was high. They showed further that the evaluation speedup could also be applied to GP image processing tasks \cite{harding2008genetic}. Langdon \cite{langdon2008simd} utilized GPUs to achieve a speedup with Tree-GP, storing full populations of trees directly on the GPU to reduce overhead. In an early application of the CUDA library for Nvidia's GPUs, Robilliard et al \cite{robilliard2009high} compared parallelization schemes and showed that GPU speedup over a CPU was problem dependent and affected by the presence of certain operators (such as the if operator). Chitty \cite{chitty2012fast} further demonstrated that the speedup effect of a GPU on GP can sometimes be exceeded by a multi-core CPU, although it is unclear whether this result would be challenged by modern graphics processing hardware.

Recent work has highlighted the capabilities of the CUDA library. Sathia et al \cite{sathia2021accelerating} showed an average speedup of 40X when comparing their CUDA accelerated GP variant against several other standard symbolic regression libraries. Truijillo et al \cite{trujillo2022gsgp} introduced the first CUDA implementation of Geometric Semantic Genetic Programming (GSGP-CUDA), showing speedups greater than 1000X. Zhang et al \cite{zhang2023gpu} successfully applied NVIDIA GPUs to accelerate GP-powered feature extraction in binary image classifiers, while Wang et al \cite{wu2026enabling} demonstrated how they could also be utilized to allow population-level parallelization options in Tree-GP. 

\section{Methods}

In this section, we discuss key aspects of the Beagle framework \cite{haut2026gpu}. Note that an earlier system called BEAGLE \cite{gagne2003distributed} is unrelated to the Beagle framework used in this work. 

\subsection{The Beagle Framework}\label{sec:beagleOverview}

Beagle is an open-source symbolic regression framework developed by Ilya Basin at Noblis, Inc., that is designed to take advantage of the massive computing power of GPUs and is specifically designed to be "friendly" to heterogeneous computing environments \cite{beagle}. One of the key advantages of this design is that it makes it possible to evaluate hundreds of millions of models in a timespan of 30 minutes or less. This evaluation efficiency can be used to either run very large population sizes or run evolution through many generations with smaller population sizes (breadth vs. depth).    
The Beagle source code can be accessed here: {\bf https://github.com/Noblis/beagle-v1.x} 


Beagle uses the ILGPU open-source C\# library. ILGPU enables programming at the lowest CUDA level (rather than through the use of higher-level GPU libraries). This, in turn, enables the implementation of arbitrary optimization techniques. Beagle’s development stack is cross-platform and can run well on both Linux and Windows operating systems.

\subsubsection{Approaches for GPU Code}

To maximize performance, Beagle can distribute work across multiple GPUs, if available. It can also handle situations where the entire population cannot fit into GPU memory at once, in which case the GPU portion is executed in multiple smaller batches that can fit. 

The GPU handles evolutionary runs and fitness function evaluations, while the CPU handles selection, the birth/death process, and mutations. In its current implementation, Beagle does not use crossover. To minimize the overhead of the resulting data exchange between GPU and CPU, which can be significant, Beagle performs multiple fitness case evaluations per individual in a single generation (with a typical batch size of 512 or 1024). For example, a single generation for a population of 1,000,000 individuals at a batch size of 512 would perform 512,000,000 total fitness case evaluations. Following this approach, Beagle never has to send the same individual to a GPU twice. In a single Beagle generation, all genomes in a population are sent to the GPU together, and each executes a batch of fitness cases. When done, the aggregate results of the fitness evaluations for every individual are returned to the CPU for the selection/birth/death/mutation processes.

When the Beagle GPU kernel is run, each CUDA block represents a single organism (i.e., the number of blocks equals the population size) and each CUDA thread in the block represents a single fitness case for that organism. This reduces, and often completely eliminates, thread diversion on the GPU, thus avoiding performance degradation related to GPU warp thread diversion. Due to NVIDIA GPU limitations, 1024 is the maximum number of threads per block, meaning 1024 is the maximum number of fitness cases allowed per individual per generation in Beagle.

\subsubsection{Approaches for CPU Code}

On the CPU side, Beagle’s implementation almost completely eliminates Garbage Collection (GC), memory fragmentation, and memory allocation/deallocation overheads, which otherwise may consume more than half of compute power for evolutionary algorithms. Usually, evolutionary algorithms need to constantly allocate and deallocate memory when individuals are born or die. In Tree-Based GP languages the problem is even more pronounced, since individual parts of the code tree are typically stored separately on the heap. This causes GC to work “overtime” and creates memory fragmentation problems, thus slowing down performance. To combat this effect, Beagle never deallocates any memory for individuals. Instead, memory from “dead” individuals is placed in a “dead pool” and recycled whenever memory is needed for a new individual. In general, Beagle never deallocates memory other than for C\# strings, thus minimizing the overhead of GC and memory defragmentation. 

Where possible, all Beagle CPU code is written to take full advantage of concurrent execution using Microsoft’s Parallels library. This concurrent execution is optimized to eliminate thread contention (i.e., no locks), leading to efficient concurrent execution across all available CPU cores. 

\subsubsection{GPU-Friendly Bespoke Language}

Beagle uses a custom-created Genome Computer Language (GCL), which is a Linear Genetic Programming (LGP) language based on Reverse Polish Notation (RPN). Using LGP, as opposed to the typical Tree-Based approach, makes the GCL more GC- and GPU-friendly because the GCL is not heap-based, resulting in better performance and scalability.

GCL is structured in a way that produces valid genomes for the majority of mutations. In cases where a valid genome is not produced directly, the mutation mechanism detects the issue and performs a random correction operation. Thus, every mutation in Beagle is guaranteed to produce a viable genome.

\subsubsection{Population Size}

The combination of the features described above enables Beagle to scale to population sizes in the millions. For example, on a relatively modest GPU server with dual Xeon Gold 6426Y and dual NVIDIA A16 chips, a generation typically takes under 0.5 seconds for a population of one million individuals using 512 fitness cases per generation. In practice, Beagle can run well even on high-end laptops with NVIDIA chips.

Once populations reach into the hundreds of thousands – and Beagle often deals with populations in the tens of millions – a ranking-based selection process starts to represent a significant bottleneck. To combat that, Beagle uses a novel high-performance Monte-Carlo-inspired population control approach that mimics a ranking approach but scales better. At a high level, this means that rather than sorting the entire population by fitness – an approach that does not scale well for large populations – Beagle sorts a random sample of 100 individuals from the population and uses this “population microcosm” to determine percentiles for the fitness distribution for the entire population. These percentiles are then used in a single-pass selection process. Although this “microcosm” approach may be somewhat imprecise, errors cancel each other out over multiple generations. This enables Beagle to efficiently maintain a target population size for large populations. For example, for a population of 1,000,000, Beagle’s approach is approximately 30,000 times faster than a traditional ranking-based approach.

\subsubsection{Fitness Functions}

Beagle supports any custom user-defined fitness function that compares fitness cases point-by-point with the target result of the training data. By default, Beagle uses a correlation-based fitness function adapted from \cite{Haut2023}. The foundation of the correlation fitness function is shown in Equation \ref{eq:corr},

\begin{equation}
\label{eq:corr}
    r = \frac{\sum_{i=1}^N (y_i - \bar{y}) (\hat{y}_i - \bar{\hat{y}})}{\sqrt{\sum_{i=1}^N (y_i - \bar{y})^2 \times \sum_{i=1}^N (\hat{y}_i - \bar{\hat{y}})^2}}
\end{equation}
where $N$ is the number of data points $i$, $y_i$ is the target output, and $\hat{y}_i$ is the model output. 

The value of $r$ is then squared. If a model produces or a dataset contains invalid numbers, rather than removing them, Beagle handles those point pairs separately and rewards or punishes models depending on whether the model invalids match the target invalids or not. Thus, $r$ is only representative of such point pairs for a model, in which both the model and the target output are valid numbers. Once $r$ is determined, it is possible to compute the total score using Equation \ref{eq:score},

\begin{equation}\label{eq:score}
    score=M r^4 (N-(c_1+c_2))-M (c_1-c_2)
\end{equation}
where $N$ is the total number of fitness cases, $M$ is a max score parameter set within Beagle, $c_1$ is the number of valid/invalid pairs, and $c_2$ is the number of invalid/invalid pairs.

In \cite{Haut2023}, evolution was performed by minimizing $1-r^2$; in Beagle, we maximize $r^4$. In Beagle we chose to square  $r^2$  so that we get a wider variance across the very good solutions. This helps prevent evolution from getting stuck at "very good" but not "perfect" models.

\subsubsection{Operator Set}\label{sec:beagleParams}

When using Beagle, we used the following operator set: $+,-,*,/$, $x^2,\sqrt{x}, 1/x$,
    $cos, sin, tan, arccos, arcsin$, 
    $arctan$, $tanh, log,$ and $exp$.


\section{Experimental Setup}





To explore the impacts of various GPU-enabled population dynamics on symbolic regression problems, we selected several problems from the Feynman Symbolic Regression Benchmark and explored each of these problems using several constant and stepped population sizes. We describe our choice of benchmark problems in Section \ref{sec:benchmark} and we describe our population setups in Section \ref{sec:popDyn}.

\subsection{Benchmark}\label{sec:benchmark}

In this contribution, we explored 7 problems from the Feynman Symbolic Regression Benchmark \cite{data}. We chose these 7 problems because in prior work we found them challenging enough that perfect solutions are not always found without being so difficult that a solution is never found. Performance in these problems is expected to be most sensitive to changes in our search process. The equations for these 7 benchmarks are shown in Equations 7, 14, 31, 56, 57, 86, 91 in Table \ref{tab:equation_list}. 

\begin{table}[h]
\centering
\caption{Selected Benchmarking Problems.}
\label{tab:equation_list}
\renewcommand{\arraystretch}{1.4}
\begin{tabular}{|c|c|c|}
\hline
Eq. & Feynman Nr. & Expression \\ \hline

(7) & I.11.19  & $x_1 y_1 + x_2 y_2 + x_3 y_3$ \\ \hline

(14) & I.13.12  & $G\,\mathrm{m1}\,\mathrm{m2}
       \left(\dfrac{1}{\mathrm{r2}}-\dfrac{1}{\mathrm{r1}}\right)$ \\ \hline

(31)  & I.30.5 & $\sin^{-1}\!\left(\dfrac{\lambda}{dn}\right)$ \\ \hline

(56) & II.6.15a  & $\dfrac{3(pz)\sqrt{x^2+y^2}}{r^5(4\pi\epsilon)}$ \\ \hline

(57) & II.6.15b  & $\dfrac{3p}{(4\pi\epsilon)\left(r^3\sin\theta\cos\theta\right)}$ \\ \hline

(86) & III.4.32  & $\dfrac{1}{\exp\!\left(\dfrac{\omega\hbar}{kT}\right)-1}$ \\ \hline

(91) & III.10.19  & $u\sqrt{\mathrm{Bx}^2+\mathrm{By}^2+\mathrm{Bz}^2}$ \\ \hline

\end{tabular}
\end{table}







For each benchmark problem, we used a training-testing split of 80-20, where 512 randomly generated points were used for training and 128 were used for testing. The success of an evolved model was determined using the test set. Specifically, we considered a problem solved if the error for all test points was less than 0.1\% (one tenth of one percent). 

For each of the experimental setups and benchmark problems, we performed 30 independent experiments and counted the number of times the best model from training passed the validation process on the test set. The number of times a search strategy led to a validated model out of the 30 trials was used to compare each of the different strategies. Each independent experiment was limited to a runtime of 15 minutes, thus evaluating the efficiency of each search strategy to find a good model in a short time frame. 

\subsection{Population Dynamics}\label{sec:popDyn}
In this work, we explore both constant population sizes and stepped population sizes. Constant population sizes were explored to observe the tradeoff between population size and generations completed under the fixed computational budget, thus allowing us to consider both very wide but shallow searches and very narrow but deep searches. Large populations allow for significant coverage of the search space, but due to the time limit, they do not get as many generations to iterate and improve. Very small populations, on the other hand, are more focused and get many generations to improve, but as a result of the small population size, may be at risk of converging to a local optimum. Since we suspect there is value in initially having a large population to identify potentially promising regions and then focusing the population on the most promising regions, we also explored several stepped strategies where we start with a large population size and then, after a set number of generations, drop down to a smaller size. 

For constant population size experiments, we explored sizes from as small as 100 individuals up to sizes as large as 100 million individuals. The scalability needed for extremely large populations was enabled by the use of a GPU-based framework. 

\section{Results}

\subsection{Population Size Stability}

Since target population sizes are not strictly enforced caps in Beagle, we wanted to explore how much variability occurs when we set a specific population size. 

\begin{figure}[h!]
\centering
\includegraphics[width=8cm]{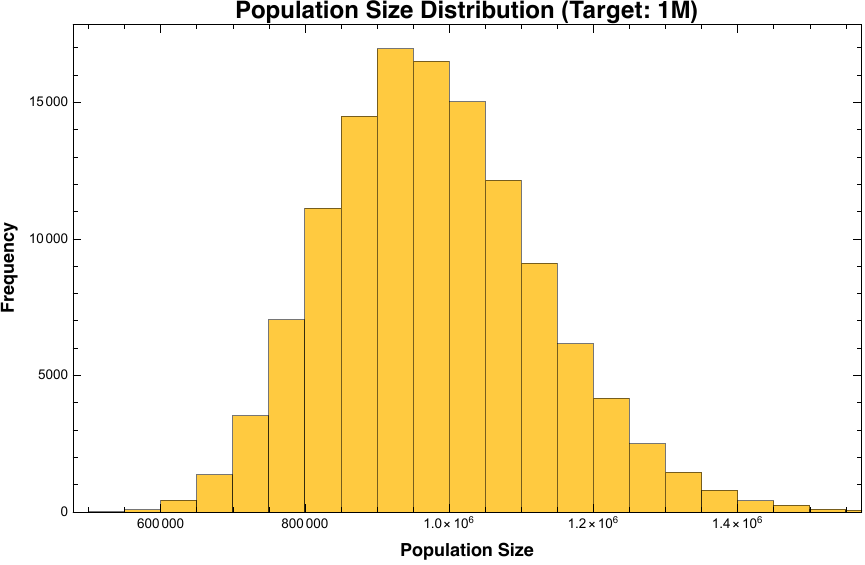}
\caption{Shown here is the distribution of actual population sizes across all generations, runs, and experiments that used a target population size of 1 million. We can see that the actual population size falls within a normal distribution around the target size.}
\label{fig:stability1m}
\end{figure}

Figure \ref{fig:stability1m} shows the distribution of population sizes across all generations for all experiments with a target population size of 1 million individuals. We can see that the target population size acts as a "soft target," resulting in the actual population size "dancing around" the designated number. From Figure \ref{fig:stability1m}, we can see that the population is distributed as a Gaussian around the target population size, with most of the population sizes within $\pm20\%$ of the target. 

\subsection{Constant Population Sizes}

\begin{figure}[h!]
\centering
\includegraphics[width=10cm]{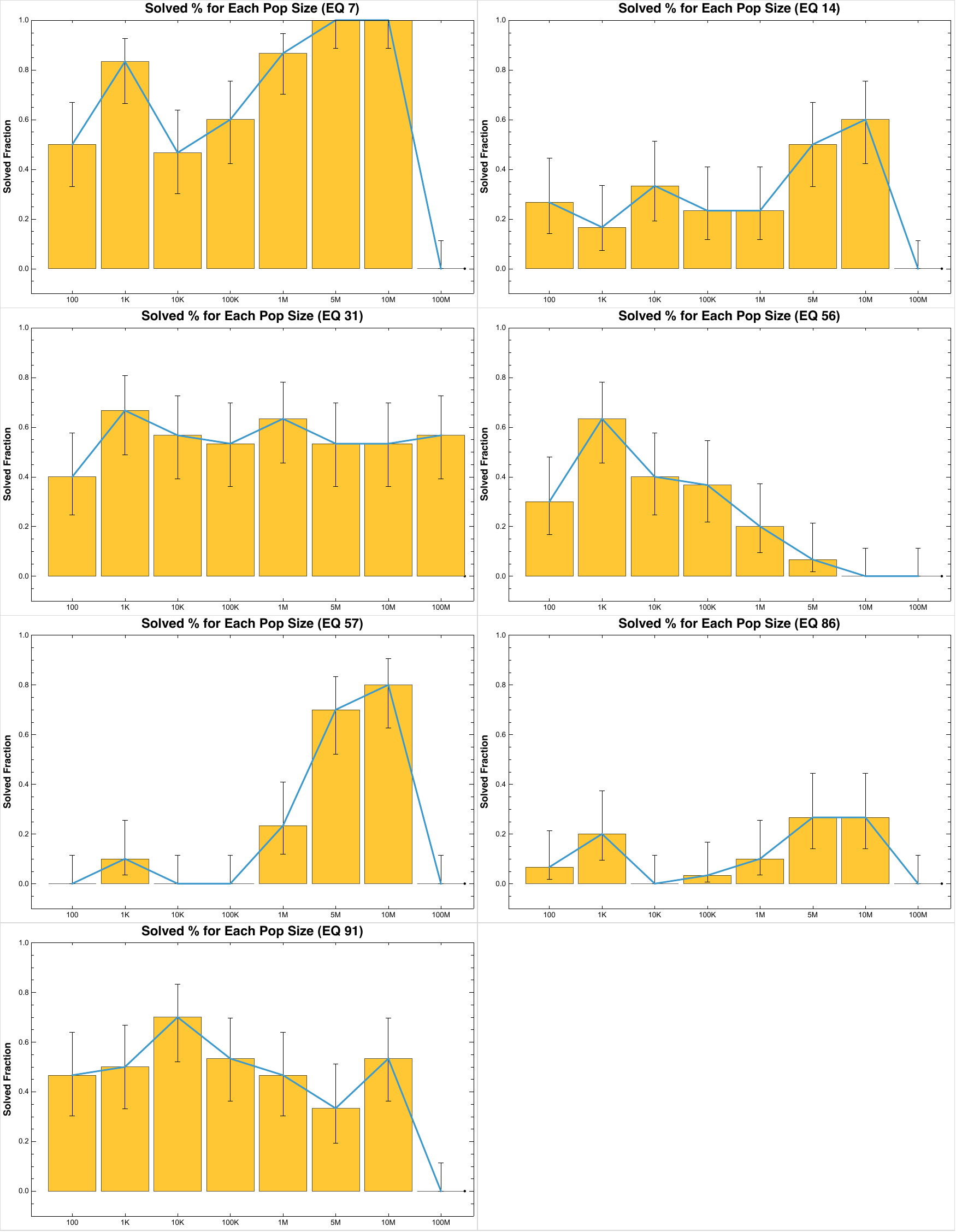}
\caption{Shown are the performance results on each equation using the different population sizes. The bar height indicates the percentage of solved problems among the 30 repeats. Confidence bands are computed using the Wilson Interval.}
\label{fig:constPerEq}
\end{figure}

The results of running constant population sizes ranging from small populations of 100 to large populations of 100 million individuals are shown in Table \ref{tab:population_vs_solved} and Figure \ref{fig:constPerEq}. Table \ref{tab:population_vs_solved} shows how many times one of the benchmark problems was solved using each population size. Figure \ref{fig:constPerEq} compares how the different populations performed for each equation. Interestingly, we see higher performance at both the high and low end (1K, 5M, and 10M), with 10M solving the most problems overall. From Figure \ref{fig:constPerEq}, we can clearly see that the best performing population size is problem dependent. For example, Equation 57 was consistently solved only when we had population sizes of 5 and 10 million. Meanwhile, Equation 56 was frequently solved with a small population of 1,000 but rarely with larger population sizes. This indicates that Equation 56 requires a lot of iteration to develop a good solution, whereas Equation 57 requires significant diversity to arrive at a good solution. We suspect that this means that the fitness landscape for finding Equation 57 is very rugged, so we need to explore a lot of the search space, whereas the fitness landscape for Equation 56 is much smoother but requires many generations to find an optimum. Future work would need to be done to confirm this. We can also see that some problems are just easier overall and are solved frequently using all population sizes (Eq 7, Eq 31, and Eq 91). 
\begin{table}[h]
\centering
\caption{Number of solved problems as a function of population size. 210 is the maximum number possible to solve if all runs produce validated solutions.}
\label{tab:population_vs_solved}
\begin{tabular}{|c|c|}
\hline
Population Size & \# Solved \\ \hline
100  & 60/210  \\ \hline
1K   & 93/210 \\  \hline
10K  & 74/210 \\  \hline
100K & 69/210 \\  \hline
1M   & 82/210 \\  \hline
5M   & 102/210 \\ \hline
10M  & 112/210 \\ \hline
100M & 17/210 \\ \hline
\end{tabular}
\end{table}

\begin{figure}[h!]
\centering
\includegraphics[width=10cm]{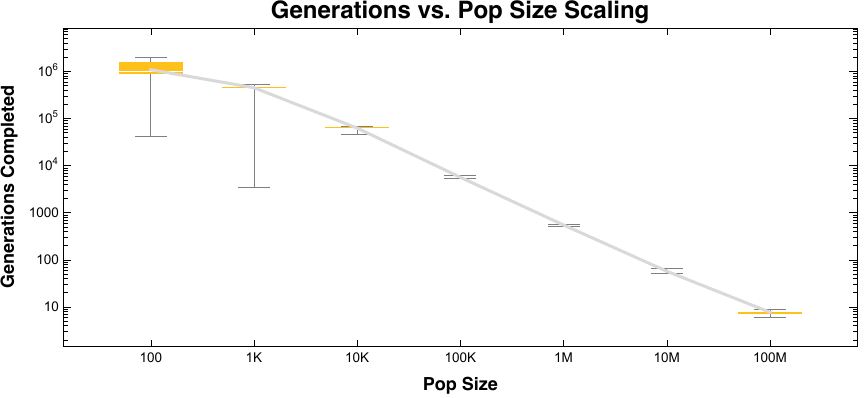}
\caption{Shown here are the number of generations completed using each population size setup for all 30 runs across the 7 benchmark problems. The results show that the scaling is approximately linear, so there is an inverse linear relationship between population size and generations.  }
\label{fig:scaling}
\end{figure}

Since all run times were constrained to 15 minutes, a natural tradeoff occurs between the number of generations completed and the population size. To see how this tradeoff scales, we aggregated the number of generations completed across all 30 runs for all 7 problems and reported the distributions for each population size. These results can be seen in Figure \ref{fig:scaling}. Since we span several orders of magnitude, the y-axis is plotted on a log-scale. We can see that the scaling between generations and population size is inverse-proportional. Thus, a 10x increase in population size leads to a 10x reduction in the number of generations completed. This balanced tradeoff confirms that Beagle has no significant computational overhead when managing population sizes up to 100 million. On the lower end of population sizes, we see the scaling degrade when using a population size of 100. 


\begin{figure}[h!]
\centering
\includegraphics[width=12cm]{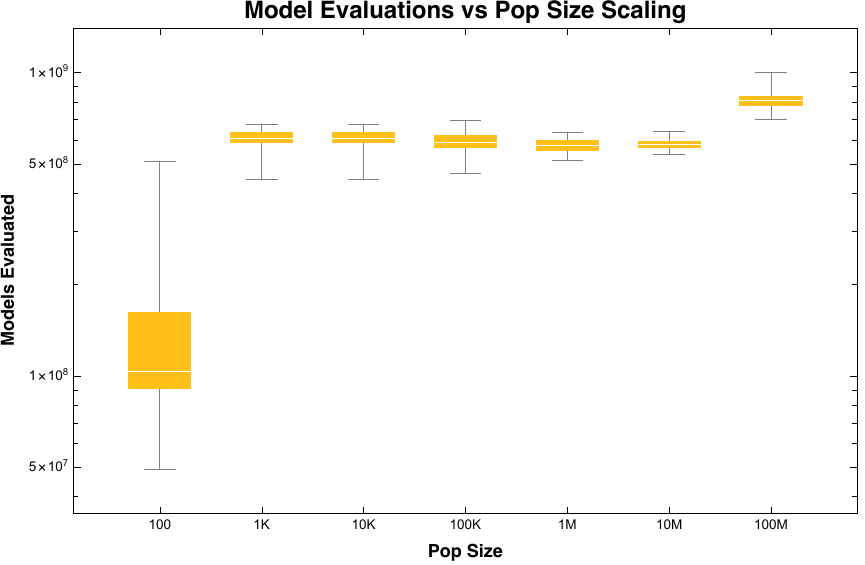}
\caption{Shown here are the total number of model evaluations over the population size setup across all 30 runs for all 7 benchmark problems. We can see that, regardless of population size, the number of model evaluations is comparable.}
\label{fig:evals}
\end{figure}

We also explored the total number of model evaluations in each of the evolutionary runs using different population sizes. The results comparing the number of evaluations completed per setup are shown in Figure \ref{fig:evals}. We can see that, for the most part, the total number of model evaluations remains about constant. However, at a population size of 100, we see significantly degraded performance. Efficiency seems to increase as we increase to 100 million, but this may be artificial as a result of how the time constraint is implemented. The time limit check occurs just once per generation. With the population size of 100 million, only 6-9 generations were completed in each run, which means that a generation may take up to 2.5 minutes to complete; if a generation starts just before the time limit, it will be allowed to be completed, effectively giving the run extra time beyond the set 15-minute limit. The effects of this "bonus" time may be negligible for smaller populations, but more significant for larger one. 

These scaling results show that the performance differences across problems for most of the setups are a reflection of the depth vs. width of the evolutionary runs and not a feature of the number of evaluations completed. An exception to this is the population size of 100, where the worse performance likely contributed to evaluating 80\% fewer models in the 15-minute time span. This worse scaling performance for the small population size is likely a result of the communication overhead incurred by frequently passing data to/from GPU while not getting much benefit from using the GPU since its massive compute power is severely underutilized for such small population.  

\subsection{Stepped Population Sizes}



\begin{figure}[h]
\centering
\includegraphics[width=12cm]{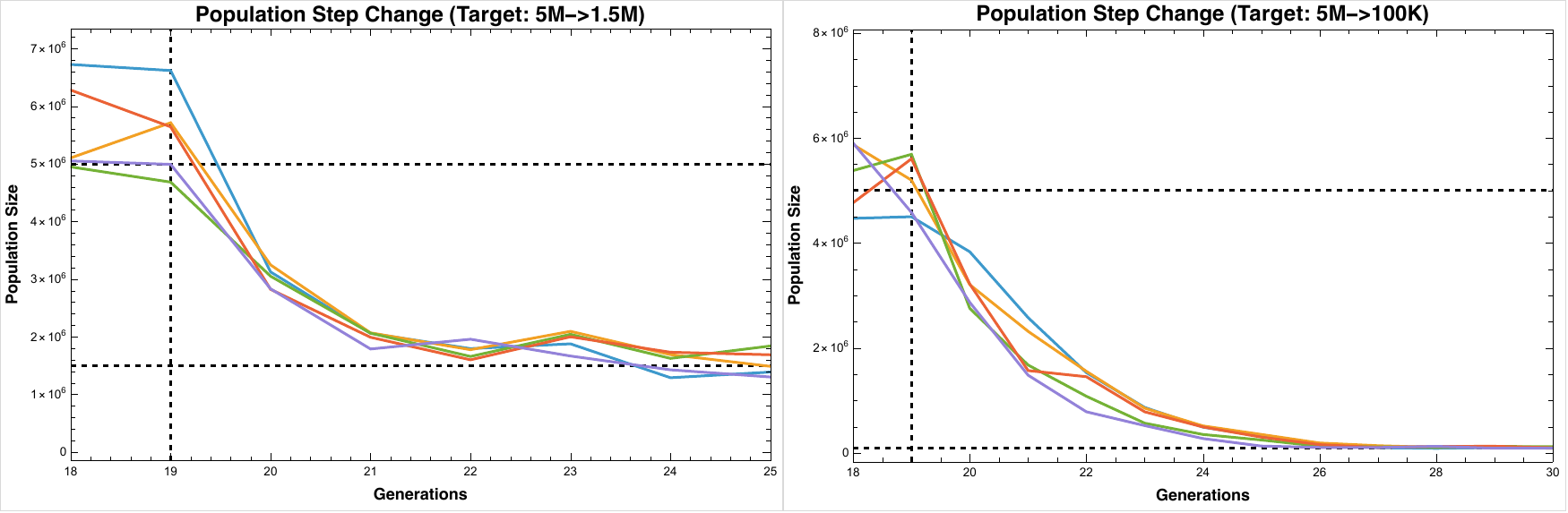}
\caption{Population size changes with a step in population size target is triggered after generation 19. Left: Step-down from 5 million to 1.5 million target size; Right: Step-down from 5 million to 100,000 target size.}
\label{fig:stepZoom}
\end{figure}

In Figure \ref{fig:stepZoom}, we explore how the population size changes immediately after a new population size is triggered. In this example, after generation 19, the target population size changes from 5 million to 1.5 million (on the left) and from 5 million to 100,000 (on the right). We can observe that the population takes 2-3 generations to reach the new target population size when going from 5 million to 1.5 million and 3-4 generations when going from 5 million to 100,000. This shows that population size changes are more of a smooth transition than a hard step in this selection approach. Still, such large changes in just 2-4 generations may be a bit extreme, so future work will explore implementing a logistic function change, which is more common in natural populations.  

\begin{figure}[h!]
\centering
\includegraphics[width=10cm]{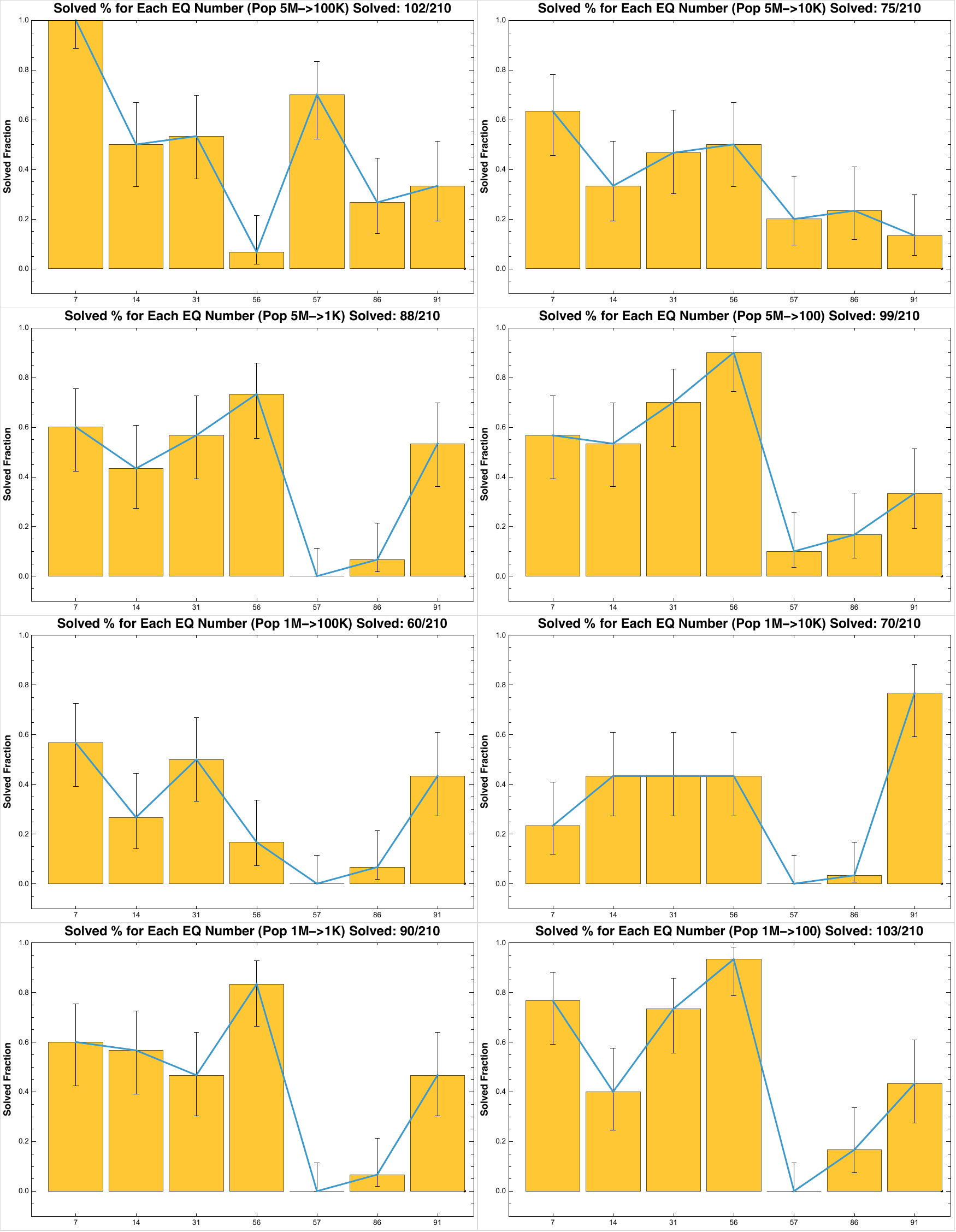}
\caption{Shown here are the performance results (solved \%) of each stepped strategy across all 7 benchmark problems. }
\label{fig:stepSummary}
\end{figure}

The results of running the stepped population sizes are shown in Figure \ref{fig:stepSummary}. It can be observed that the 5 million to 100,000 step and the 1 million to 100 step perform best (solving the most problems). The performance of the 5 million to 100,000 setup is likely preferred since this setup at least occasionally solves all of the problems, whereas none of the 1 million starting size setups ever solve Equation 57. It is interesting to note the difference in performance on Equation 56, since in the 5M->100K setup the problem is rarely solved, while in the 1M->100 case it is nearly always solved. This observation aligns with the results from the constant population size exploration that indicated Equation 56 requires many generations to be solved. While the 5 million to 100 thousand setup led to the most overall solutions, the 5 million to 10 thousand setup is interesting because it seemed to have the best balance in performance across all problems, with the worst problem-specific success rate being around 20\%. 

\section{Conclusions}

In this contribution, we used the GPU capabilities of the Beagle framework to explore the impact that population dynamics have on the success of solving various challenging symbolic regression problems. The utilization of GPUs allows us to evaluate around 600 million models in each 15-minute run so we could explore very wide populations with few generations (Example: 5 million individuals through 100 generations) along with narrow populations through many generations (Example: 1,000 individuals through 600,000 generations). 

Although GPU technology keeps developing at a rapid pace and our results are therefore a snapshot of today's capabilities, they demonstrate that the time is ripe for seriously scaling GP runs to large population sizes. It is anticipated that this capability will make some difficult problems accessible to solutions with GP for the very first time. 

Beagle is open-source and we encourage everyone in need of more GP power to experiment with GPU-enabled frameworks.

\begin{credits}
\subsubsection{\ackname}    The team acknowledges Noblis, Inc. for supporting the development of the Beagle framework. This work was supported in part by Michigan State University through computational resources provided by the Institute for Cyber-Enabled Research.

\end{credits}
%
%
%
%
\clearpage
\bibliographystyle{abbrv}
\bibliography{references}

\end{document}